\documentclass[11pt]{article}

\usepackage[final]{acl}

\usepackage{times}
\usepackage{latexsym}

\usepackage[T1]{fontenc}

\usepackage[utf8]{inputenc}

\usepackage{microtype}

\usepackage{inconsolata}

\usepackage{graphicx}
\usepackage{mathtools}
\usepackage{amssymb}   
\usepackage{amsmath}  
\usepackage{booktabs}
\usepackage{multirow}
\usepackage{todonotes}

\usepackage[table]{xcolor}

\title{{DiCALM: Converting Autoregressive Convolution-Attention Language Models into Discrete Diffusion Language Models}}

\author{Phong Le \and Ngoc-Quan Pham \and Dat Quoc Nguyen \\
        Qualcomm AI Research\thanks{Qualcomm Vietnam Company Limited. Qualcomm AI Research is an initiative of Qualcomm Technologies, Inc.} \\
        \texttt{\{phonle, ngocquan, datnq\}@qti.qualcomm.com}}

\begin{document}
\maketitle
\begin{abstract}
Existing methods for adapting autoregressive (AR) language models to diffusion generation focus mainly on full-attention architectures. More efficient hybrid architectures remain underexplored. We introduce DiCALM, a recipe for converting the hybrid convolution–attention models of the LFM2 family (LFM2 and LFM2.5) from AR to masked-diffusion generation. DiCALM progressively extends LFM2's causal gated convolution so that it captures local context in both directions. It combines this extension with a masking warm-up and joint AR--diffusion training. In controlled experiments, DiCALM performs comparably to the corresponding fine-tuned LFM2.5-based AR models and delivers consistent speedups over them at both the 350M and 1.2B scales: over $1.4\times$ on average and up to $1.79\times$.
\end{abstract}

\begin{figure*}[!t]
  \centering
  \includegraphics[width=\linewidth]{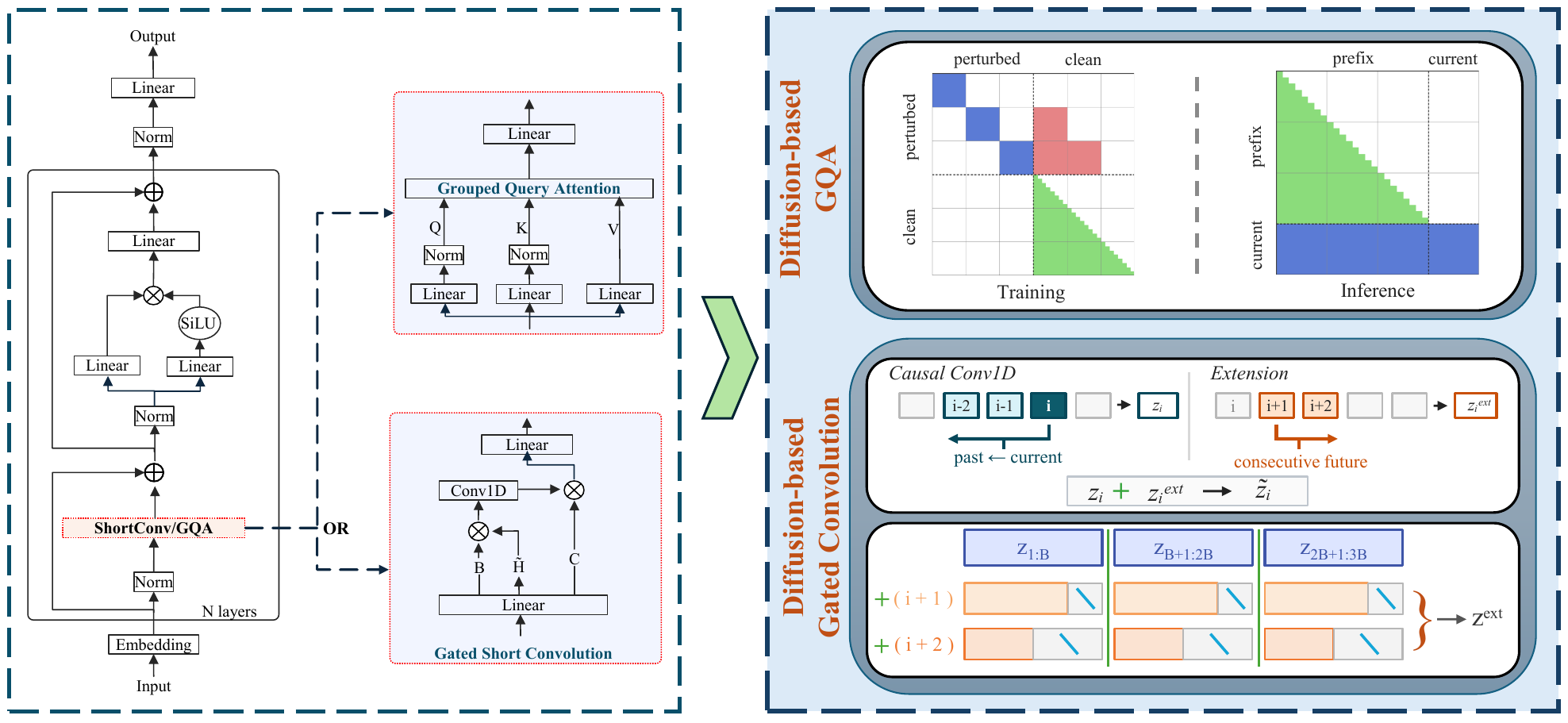}
  \caption{\textit{Left}: Architecture of the LFM2 model family. \textit{Right}: Applying DiCALM to LFM2, with mask design for the grouped-query attention layers and a bidirectional extension of the gated short convolution layers.}
  \label{fig:overview}
\end{figure*}

\section{Introduction}

Recently, diffusion language models based on masked prediction (MDLMs)~\cite{mdlm, llada, blockdiffusion, dream7b, llada2} have gained attention as an alternative~\cite{llada2, nemotronlabsdiffusion} to standard autoregressive (AR) language models. Unlike AR models, which generate tokens one at a time from left to right, MDLMs predict multiple masked positions at once in each decoding step. The number of denoising steps can be adjusted too: fewer steps mean faster decoding but lower generation quality, giving a flexible quality--speed trade-off~\cite{llada, dream7b}.

Given the computational challenge of scaling MDLMs from scratch, recent works instead adapt AR-initialized models~\cite{dream7b, llada2, fastdllmv2, nemotronlabsdiffusion}, by introducing attention warmup~\cite{llada2}, AR-like token-shift~\cite{dream7b, fastdllmv2, idlm}, block-diffusion training~\cite{fastdllmv2, sdar, llada2}, and joint AR-dLM objectives~\cite{nemotronlabsdiffusion, idlm, flare}. However, current studies focus mainly on full-attention architectures, while the transfer of hybrid models~\cite{nemotron3, lfm2, qwen35blog} remains underexplored~\cite{flare}. 
By combining quadratic-complexity attention~\cite{vaswani2017attention, gqa} with efficient sequence-mixing modules~\cite{mamba2, mamba3, gla, gateddeltanet}, hybrid architectures offer a better balance of accuracy and inference efficiency, yet these combinations make the existing direct conversion recipes less straightforward.

In this context, the LFM2 family---including LFM2 and LFM2.5~\cite{lfm2, liquidAI20261p2B}---stands out as a hybrid model family designed specifically for on-device deployment. It introduces gated short convolution (ShortConv) alongside a small number of grouped-query attention (GQA) layers~\cite{gqa}. This architecture currently delivers models having the fastest inference among similarly sized alternatives on CPUs and GPUs with a low memory footprint~\cite{liquidAI20261p2B, liquidAI2026350M}. This motivates us to convert models in the LFM2 family into discrete diffusion language models to further accelerate inference while preserving model quality. Our focus on LFM2 differs from that of the concurrent work FLARE \cite{flare}, which applies diffusion conversion to Qwen3.5 \cite{qwen35blog}. Because the two works target different architectures, we do not include a direct comparison with FLARE in this paper.

To this end, we present \textbf{Di}ffusion \textbf{C}onvolution--\textbf{A}ttention \textbf{L}anguage \textbf{M}odel (DiCALM), a novel adaptation recipe for converting LFM2's hybrid ShortConv-GQA architecture from an autoregressive language model into a masked diffusion language model. DiCALM progressively introduces bidirectional local context into LFM2's causal gated ShortConv layers with minimal changes to their one-dimensional convolution operations. It combines these architectural changes with masking warm-up and joint AR-MDLM training to transition the pretrained model from autoregressive prediction to bidirectional masked prediction. 
Under controlled conditions with the same initialization and training data, DiCALM achieves performance competitive with an equivalently fine-tuned LFM2.5-based AR model, while delivering average decoding speedups of 1.43× (1.2B parameters) and 1.42× (350M parameters). Peak speedups reach 1.73× and 1.79×, respectively.

\section{Our DiCALM}
The section describes the model architecture and the training recipe for DiCALM.
\setlength{\belowdisplayskip}{7pt} \setlength{\belowdisplayshortskip}{7pt}
\setlength{\abovedisplayskip}{-1pt} \setlength{\abovedisplayshortskip}{-1pt}

\subsection{Model Architecture}

Figure~\ref{fig:overview} presents our DiCALM hybrid architecture, which integrates diffusion-based gated convolution with diffusion-based GQA layers. Building on the LFM2 family~\citep{lfm2}, we convert its causal gated short convolution and GQA layers into blockwise attention with causal context. This design enables bidirectional denoising within the current block while preserving causal attention for prompt prefilling and KV-cache updates. %

\subsubsection{Diffusion-based Gated Convolution}\label{para: ShortConvX}

The operations of the LFM2's causal gated short convolution \citep{lfm2} are formulated as: 

{\small 
\begin{subequations}
\label{eq:short-conv-gate}
\begin{align}
(\mathbf{B}, \tilde{\mathbf{H}}, \mathbf{C}&)\text{\ }\text{\ }\text{\ } = \mathrm{Linear}_{\mathrm{in}}(\mathbf{H}) \\
 &\mathbf{\text{\ }Y}= \mathbf{B} \odot \tilde{\mathbf{H}} \\
    &\left.\begin{aligned}
      \mathbf{z}_i &= \sum_{j=1}^{k} \mathbf{w}_j \odot \mathbf{Y}_{:,i-k+j} \\
      \mathbf{Z}   &= [\mathbf{z}_1, \mathbf{z}_2, \dots, \mathbf{z}_n]
  \end{aligned}\right\} \substack{\text{\small Causal} \\ \text{\small depthwise} \\ \text{\small Conv1D}} \\
  &\mathbf{\text{\ }O} = \mathrm{Linear}_{\mathrm{out}}(\mathbf{C} \odot \mathbf{Z})
\end{align}
\end{subequations}
}

\noindent where $\mathbf{B}, \tilde{\mathbf{H}}, \mathbf{C}, \mathbf{H} \in \mathbb{R}^{d \times n}$ with $n$ being the input sequence length; $\mathrm{Linear}_{\mathrm{in}} : \mathbb{R}^{d \times n} \to \mathbb{R}^{3d \times n}$ is the upward linear projection for the input hidden states; $\odot$ denotes element-wise multiplication; $\mathbf{w}_1, \mathbf{w}_2, \dots, \mathbf{w}_k \in \mathbb{R}^{d}$ are the per-offset weight vectors of a depthwise 1-D convolution with kernel size  $k$ (padding is not shown for simplification); and $\mathrm{Linear}_{\mathrm{out}} : \mathbb{R}^{d \times n} \to \mathbb{R}^{d \times n}$ denotes linear projection from intermediate states to the final output. 

Within the causal gated short-convolution layers, the causal depthwise Conv1D is the only cross-position operator.  Converting Conv1D into a dLM formulation therefore requires extending its convolution to also incorporate future positions. This extension is nontrivial: it must (i) preserve the priors encoded in the well-trained causal kernels and (ii) introduce minimal additional computation. To this end, we extend the causal depthwise Conv1D with an auxiliary forward-looking branch that attends to $k'$ subsequent positions via element-wise operations over the corresponding future tokens. The resulting convolution is defined as follows:

\begin{subequations}
\label{eq:diff-gated-conv}
\begin{align}
  \mathbf{z}^{\mathrm{ext}}_i
    &= \sum_{j=1}^{k'} m_{i,j}\;\mathbf{w}'_{j}\odot\mathbf{Y}_{:,i+j}
    \label{eq:conv-extend-masked} \\
  \tilde{\mathbf{z}}_i
    &= \mathbf{z}_i + \mathbf{z}^{\mathrm{ext}}_i \label{eq:conv-bidir-masked} \\
    \mathbf{Z}   &= [\tilde{\mathbf{z}}_1, \tilde{\mathbf{z}}_2, \dots, \tilde{\mathbf{z}}_n]
\end{align}
\end{subequations}
\noindent where $\mathbf{w}'_1, \mathbf{w}'_2, \dots, \mathbf{w}'_{k'} \in \mathbb{R}^{d}$ are additional per-offset weight vectors; and $m_{i,j} \in \{0,1\}$ is a position-dependent mask that encodes structural constraints, e.g. block-boundary masking in blockwise diffusion~\cite{blockdiffusion}. Our extension is post-hoc and drop-in, enabling flexible switching between causal and bidirectional processing.

\subsubsection{Causal GQA to Diffusion-based GQA}

Blockwise diffusion models~\citep{blockdiffusion} operate by appending new blocks of \textsc{[mask]} tokens into the current decoded sequence and iteratively unmasking positions in the blocks. 

When the current block is fully unmasked, the KV cache is updated and a new masked block is appended; this repeats until a stopping criterion is met (the end-of-sentence token unlocked or reaching the maximum length). 
Adapting causal GQA layers to blockwise diffusion primarily involves modifying the attention mask, a design choice explored extensively in prior work \citep{dream7b, fastdllmv2, sdar, nemotronlabsdiffusion}. Following \citet{nemotronlabsdiffusion}, we apply bidirectional attention within the current decoding block and causal attention over the preceding context, which differs from the conventional block diffusion~\cite{blockdiffusion} that also uses bidirectional attention for its preceding blocks of context. 
Figure~\ref{fig:overview} illustrates the attention mask used by Diffusion-based GQA during training and inference. Training blockwise diffusion models requires a specialized mask that enables the loss for each randomly masked block to be computed in parallel, conditioned on the preceding clean context.

\subsection{Training Recipe}

\paragraph{Joint AR--diffusion training.} Inspired by recent dLM adaptation work~\cite{samuel2026dualobjective, nemotronlabsdiffusion}, training a block-wise diffusion model can be efficiently implemented with double-length inputs and a joint AR--dLM training objective.  Each input is formed by concatenating a randomly masked sequence (the \emph{perturbed} side) with its originally unmasked counterpart (the \emph{clean} side) -- more details can be found in Appendix~\ref{app:recipe_moredetail}. Our  training loss is defined as follows:

\begin{subequations}
\scriptsize
\setlength{\jot}{0pt}
\label{eq:loss}
\begin{align}
\mathcal{L}(\theta; \mathbf{x})
&=
(1-\alpha)\,\tau\,\mathcal{L}_{\mathrm{AR}}(\theta; \mathbf{x})
+
\alpha\,\mathcal{L}_{\mathrm{diff}}(\theta; \mathbf{x})
\label{eq:loss-total}
\\[2pt]
\mathcal{L}_{\mathrm{AR}}(\theta; \mathbf{x})
&=
-\frac{1}{|\mathbf{x}|}\sum_i \log \mathrm{P}_{\theta}\!\left( x_i \mid \mathbf{x}_{<i} \right)
\label{eq:loss-ar}
\\[2pt]
\mathcal{L}_{\mathrm{diff}}(\theta; \mathbf{x})
&=
\frac{1}{\sum_b|\mathcal{M}_{t_b}^{b}|}
\sum_{b=1}^{B}
\sum_{i\in\mathcal{M}_{t_b}^{b}}
\Biggl(
-\log \mathrm{P}_{\theta}\!\left(
[\mathbf{x}^{b}]_i \,\middle|\, \mathbf{x}_{t_b}^{b}, \mathbf{x}^{<b}
\right)
\notag\\
&\quad\quad
{}+
\max\!\Bigl\{
0,\,
\log \rho
-
\log\!\left(
\mathrm{P}_{\theta}\!\left(
[\mathbf{x}^{b}]_i \,\middle|\, \mathbf{x}_{t_b}^{b}, \mathbf{x}^{<b}
\right)
\right)
\Bigr\}
\Biggr)
\label{eq:loss-diff}
\end{align}
\end{subequations}

\noindent Here, $\mathbf{x}$ is a clean sequence, with $x_i$ denoting the token at position $i$ and $\mathbf{x}_{<i}$ denoting the sequence of all preceding tokens. We partition $\mathbf{x}$ into $B$ contiguous blocks, where $\mathbf{x}^{b}$ denoting the $b$-th clean block and $\mathbf{x}^{<b}$ denoting the sequence of all preceding clean blocks. Each block is then independently corrupted by the forward noising process at a noise level $t_b\sim\mathcal{U}[0,1]$ -- i.e., the fraction of tokens masked in block $b$ -- yielding the perturbed block $\mathbf{x}_{t_b}^{b}$. Also, $\mathcal{M}_{t_b}^{b}$ denotes the set of masked positions within block $b$, and $[\mathbf{x}^{b}]_i$ for the clean target token at position $i\in\mathcal{M}_{t_b}^{b}$.

The AR loss $\mathcal{L}_{\mathrm{AR}}(\theta; \mathbf{x})$ is the cross-entropy loss for next-token prediction on the clean input, using causal GQA and the causal gated short convolution---that is, our diffusion-based gated convolution formulation without the auxiliary forward-looking branch, obtained by setting $\tilde{\mathbf{z}}_i = \mathbf{z}_i$ instead of $\tilde{\mathbf{z}}_i = \mathbf{z}_i + \mathbf{z}^{\mathrm{ext}}_i$ (Equation~\ref{eq:conv-bidir-masked}).

The masked diffusion loss $\mathcal{L}_{\mathrm{diff}}(\theta; \mathbf{x})$ applies cross entropy only at masked positions, using blockwise GQA and diffusion-based gated convolution. We further add a regularization term to $\mathcal{L}_{\mathrm{diff}}$ with a margin that activates whenever the ground-truth probability $\mathrm{P}_{\theta}\left([\mathbf{x}^{b}]_i,\middle|,\mathbf{x}_{t_b}^{b},\mathbf{x}^{<b}\right)$ falls below a predefined threshold $\rho$. This pushes the model not merely to rank the ground-truth token first, but to assign it a sufficiently high probability -- a property that also facilitates threshold-based decoding~\cite{fastdllm}, in which a \textsc{[MASK]} token is unmasked once the probability of its top candidate exceeds a predefined threshold. We also adapt the token-shift technique~\cite{dream7b, fastdllmv2, idlm}, which predicts each masked token from the logit at its preceding position, to better preserve the representational properties inherited from AR pretraining. 

Because the model is initialized from an AR checkpoint and the diffusion side operates on noised inputs,  $\mathcal{L}_{\mathrm{AR}}(\theta; \mathbf{x})$ and $\mathcal{L}_{\mathrm{diff}}(\theta; \mathbf{x})$ differ substantially in magnitude. Following \citet{idlm}, we rescale the AR loss by a gradient-detached factor $\tau = \mathcal{L}_{\mathrm{diff}}(\theta; \mathbf{x})/\mathcal{L}_{\mathrm{AR}}(\theta; \mathbf{x})$ to bring the two terms onto a comparable scale, and use a mixing coefficient $\alpha$ to balance them. To stabilize early training, we linearly warm up $\alpha$ over the first 5\% of training steps and then hold it at its target value for the remainder of training.

\paragraph{Masking Warm-up for GQA layers.}

\begin{figure}[!t]
  \centering
  \includegraphics[width=0.95\linewidth]{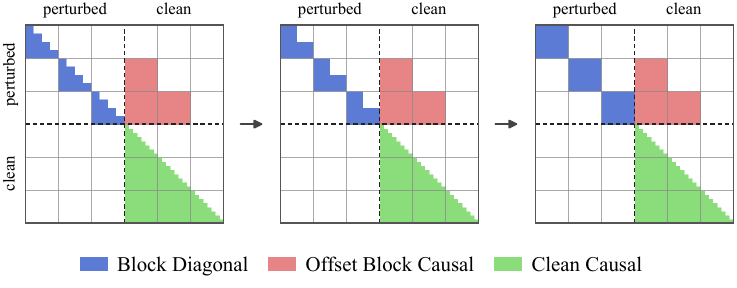}
  \caption{Our masking warm-up strategy.}
  \label{fig:masking}
\end{figure}

Inspired by~\cite{llada2}, we initialize the GQA attention masks using a near-causal sub-blockwise scheme and gradually increase the sub-block size during the second half of the warm-up steps, as illustrated in Figure~\ref{fig:masking}. We use only sub-block sizes that evenly divide the block size, ensuring that each block contains complete sub-blocks. A square-root schedule is applied for smoother sub-block transitions. The masking warm-up is applied along with the warm-up of $\alpha$ and $\rho$ (in Equations \ref{eq:loss-total} and \ref{eq:loss-diff}) with their corresponding schedules.

\section{Experiments}

\subsection{Setup}

\textbf{Model initialization \& Training configurations.}\ \ We evaluate DiCALM against AR baselines on the 1.2B and 350M scales. At each scale, DiCALM and its AR counterpart start from the same base model and are supervised fine-tuned on the same dataset. We use LFM2.5-1.2B-Base and LFM2.5-350M-Base as the base models; these use the LFM2 architecture but are pretrained on a larger corpus than the previous pre-trained LFM2 models.

We extract and deduplicate prompts from Nemotron-Cascade-SFT-Stage-2~\cite{cascadestage2}, then regenerate the responses with Qwen3-Next-80B-A3B-Instruct for consistency. We focus on general instruction-following capabilities and exclude explicit long-thinking mode traces marked by \texttt{<think>...</think>}. The resulting dataset contains approximately 2M samples, each with a maximum length of 4,096 tokens.

To ensure a fair comparison, we use the same fine-tuning setup for all models, with the learning rate as the only setting that varies across scales. For DiCALM, we use a block size of 32 and extend the gated convolution to cover up to two future positions. Full training and transfer settings are provided in Appendix~\ref{app:TransferConfig}.

\textbf{Inference.} To avoid unmasking isolated high-confidence positions, which may degrade generation quality, we apply threshold-based decoding~\cite{fastdllm, fastdllmv2} with the PARD strategy~\cite{pard}. Specifically, at each step DiCALM unmasks the longest leftmost sequence of masked positions whose confidence exceeds the threshold, leaving the remaining positions masked for subsequent steps. If no position meets the threshold, only the leftmost masked position is unmasked. Across all experiments, we use a block size of 32, a sampling temperature of 0.0, and a confidence threshold of 0.9.

\textbf{Benchmarks.} We evaluate model performance using the `lm-evaluation-harness' library and measure decoding speed separately.  Full settings are provided in Appendix~\ref{app:benchmark}.

\subsection{Main Results}

\begin{table*}[!t]
\centering
\resizebox{\textwidth}{!}{%
\setlength{\tabcolsep}{5pt}
\renewcommand{\arraystretch}{1.12}
\begin{tabular}{@{}clccccccc@{}}
\toprule
\textbf{Scale} & \textbf{Model} &
\textbf{MATH500} & \textbf{GSM8K} & \textbf{IFBench} & \textbf{BBH} &
\textbf{GPQA-Diamond} & \textbf{ARC-Challenge} & \textbf{MMLU-Pro} \\
& & \footnotesize(Flex. EM) & \footnotesize(Flex. EM) & \footnotesize(Avg. of 4) &
\footnotesize(Flex. EM) & \footnotesize(Flex. EM) & \footnotesize(EM) & \footnotesize(Flex. EM) \\
\midrule
\multirow{3}{*}{1.2B} & LFM-SFT & 22.00 & 62.55 & 20.29 & 28.28 & 14.65 & 62.12 & 16.12 \\
& DiCALM & 21.20 & 60.73 & 18.78 & 27.00 & 20.20 & 66.04 & 16.61 \\
\rowcolor{blue!6}
& \textit{TPF (speedup)} & 1.87 (1.73$\times$) & 1.64 (1.56$\times$) & 1.60 (1.52$\times$) &
1.60 (1.48$\times$) & 1.48 (1.32$\times$) & 1.28 (1.20$\times$) & 1.33 (1.17$\times$) \\
\midrule
\multirow{3}{*}{350M} & LFM-SFT & 11.80 & 21.99 & 14.67 & 22.38 & 15.15 & 25.43 & 4.31 \\
& DiCALM & 11.00 & 23.65 & 13.91 & 22.42 & 18.69 & 45.48 & 4.20 \\
\rowcolor{blue!6}
& \textit{TPF (speedup)} & 1.62 (1.53$\times$) & 1.40 (1.33$\times$) & 1.87 (1.79$\times$) &
1.58 (1.49$\times$) & 1.63 (1.49$\times$) & 1.17 (1.11$\times$) & 1.37 (1.21$\times$) \\
\bottomrule
\end{tabular}
}
\caption{Benchmark performance and inference efficiency of DiCALM and LFM-SFT at the 1.2B and 350M parameter scales. \textbf{LFM-SFT}: autoregressive baselines fine-tuned from LFM2.5-1.2B-Base and LFM2.5-350M-Base, respectively. \textbf{DiCALM}: our method evaluated using diffusion decoding with a confidence threshold of 0.9, achieving average throughput speedups of 1.43× (1.2B) and 1.42× (350M) over LFM-SFT. Column headers specify the benchmark metrics. Each \textit{TPF (speedup)} entry gives DiCALM's tokens per forward pass and, in parentheses, its throughput speedup over LFM-SFT, which generates one token per forward pass.}
\label{tab:DiCALM-main-exp}
\end{table*}

Table~\ref{tab:DiCALM-main-exp} compares DiCALM against its autoregressive baseline, LFM-SFT, at the 1.2B and 350M scales across seven benchmarks. 

At 1.2B, DiCALM largely preserves the decoding quality of LFM-SFT, and in several cases exceeds it. On generation-heavy benchmarks (MATH500, GSM8K, IFBench, and BBH), the accuracy reductions are small, bounded within $1.82$ points (e.g., $22.00 \rightarrow 21.20$ on MATH500 and $62.55 \rightarrow 60.73$ on GSM8K), suggesting that greater decoding parallelism adds only mild sensitivity in long-form generation. More strikingly, DiCALM matches or \emph{surpasses} LFM-SFT on multiple-choice benchmarks: it improves GPQA-Diamond from $14.65$ to $20.20$, ARC-Challenge from $62.12$ to $66.04$, and MMLU-Pro from $16.12$ to $16.61$. 
At 350M, score drops on generation-heavy tasks are at most $0.80$ points (e.g., $11.80 \rightarrow 11.00$ on MATH500). Multiple-choice scores improve on GPQA-Diamond and ARC-Challenge and decrease by only $0.11$ points on MMLU-Pro. This consistency across scales suggests that DiCALM does not systematically degrade quality; instead, its diffusion-based decoding retains---and on short-answer tasks can even sharpen---the original capabilities of the baseline.

In addition, DiCALM consistently reduces inference latency, with speedups from $1.17\times$ to $1.73\times$ (averaging $1.43\times$) at 1.2B and from $1.11\times$ to $1.79\times$ (averaging $1.42\times$) at 350M. At 1.2B, the gains are largest on generation-heavy tasks, which reach the highest tokens-per-forward (TPF) values of $1.60$--$1.87$ and speedups of $1.48\times$--$1.73\times$, as their longer responses let more tokens be decoded in parallel. Multiple-choice benchmarks yield smaller speedups of $1.17\times$--$1.32\times$: their long inputs and short outputs make prompt processing a larger share of inference time, leaving less to gain from diffusion decoding. This is clearest on MMLU-Pro, where a TPF of $1.33$ yields a speedup of only $1.17\times$. A similar trend holds at 350M, where gains are generally larger on generation-heavy tasks than on multiple-choice tasks. GPQA-Diamond is the exception among multiple-choice benchmark tasks, reaching $1.49\times$ speedup alongside a relatively high TPF of $1.63$.

Overall, DiCALM offers larger efficiency gains on generation-heavy workloads and smaller but consistent gains on short-answer tasks, without a consistent loss in accuracy. This favorable quality--speed tradeoff confirms the effectiveness of DiCALM in accelerating the baseline while largely retaining its original capabilities.

\subsection{Ablation Studies}

\paragraph{Component Ablations.} We ablate Convolution Extension, Margin Regularization, and Masking Warm-up in the 350M model to evaluate their contributions to IFBench performance. Each variant is retrained with one component removed and all other training settings unchanged from the full DiCALM model. Table~\ref{tab:DiCALM-ablation-components} shows that removing any component lowers the IFBench score, with the largest drop of $1.88$ points occurring when Margin Regularization is removed. These results support the contribution of each component to DiCALM’s IFBench performance.

\begin{table}[t]
\centering
\small
\setlength{\tabcolsep}{6pt}
\renewcommand{\arraystretch}{1.08}
\begin{tabular}{@{}ll}
\toprule
\textbf{Setting} & \textbf{IFBench (Avg. of 4)} \\
\midrule
\textbf{DiCALM (full)} & 13.91 \\
w/o Convolution Extension & 12.97 {\textcolor{red}{(-0.94)}} \\
w/o Margin Regularization & 12.03 {\textcolor{red}{(-1.88)}} \\
w/o Masking Warm-up & 13.30 {\textcolor{red}{(-0.61)}} \\
\bottomrule
\end{tabular}
\caption{Component ablations on IFBench for the DiCALM 350M model variant. Each ablation removes one component at a time while keeping the others unchanged. Red parenthetical values show the score-point decrease relative to the full DiCALM setting.}
\label{tab:DiCALM-ablation-components}
\end{table}

\paragraph{Decoding-Threshold Trade-off} The threshold sweep for the 1.2B model on IFBench (Table~\ref{tab:DiCALM-threshold-ablation}) shows a clear quality–speed trade-off. Lowering the confidence threshold from 0.9 to 0.5 increases the speedup from 1.52× to 2.18× but reduces the IFBench score from $18.78$ to $15.45$. The confidence threshold of 0.9 balances quality and speed.

\begin{table}[t]
\centering
\small
\setlength{\tabcolsep}{5pt}
\renewcommand{\arraystretch}{1.08}
\begin{tabular}{@{}cccc@{}}
\toprule
\textbf{Threshold} & \textbf{IFBench score} & \textbf{TPF} & \textbf{Speedup} \\
\midrule
0.9 & 18.78 & 1.60 & 1.52$\times$ \\
0.7 & 16.16 & 1.83 & 1.74$\times$ \\
0.5 & 15.45 & 2.25 & 2.18$\times$ \\
\bottomrule
\end{tabular}
\caption{IFBench results for DiCALM 1.2B at decoding confidence thresholds of 0.9, 0.7, and 0.5.}
\label{tab:DiCALM-threshold-ablation}
\end{table}

\section{Conclusion}

We introduced DiCALM, a recipe for converting models in the LFM2 family into masked diffusion language models through a bidirectional convolution extension, joint AR–diffusion training, and masking warm-up. At the 1.2B and 350M parameter scales, DiCALM accelerates decoding while maintaining performance comparable to equivalently fine-tuned LFM2.5-based AR models. These results highlight the potential of masked diffusion language modeling for hybrid architectures.

\section*{Limitations}

We restrict the conversion to supervised fine-tuning; extending DiCALM across multiple training phases---such as continual pretraining followed by SFT---could further strengthen its capabilities. Finally, following prior efforts, this work targets general instruction-following. We therefore exclude explicit long thinking-mode data and leave its integration to future work. Future work could also explore extending DiCALM to other convolution-based architectures and improving its training and decoding strategies.

\bibliography{references}

\newpage 

\appendix

\section{Training and Transfer Details}

\subsection{More details of our training recipe} \label{app:recipe_moredetail}

For data processing, we employ sequence packing based on~\cite{fastdllmv2}, but avoid fragmenting training samples in order to mitigate minibatch-size variance. Raw samples are first right-padded with \textsc{[eos]} tokens so that their lengths become divisible by the target block size. These \textsc{[eos]}-padded samples are then iteratively appended to form clean training sequences until the defined maximum training length is reached. Any remaining space that none of the available \textsc{[eos]}-padded samples can fit is filled with \textsc{[pad]} tokens.

Along with sequence packing, we construct special attention masks (i.e., document masks) for the packed samples to prevent data leakage across samples. Since all raw samples are properly right-padded with \textsc{[eos]} tokens from the start, these document masks integrate naturally with the blockwise diffusion mask without affecting the overall blockwise structure. 

For the training of the Diffusion-based Gated Convolution, we build the required blockwise indices and gather the corresponding values to condition each perturbed block on its clean context. Moreover, because these blockwise indices are fixed to a predefined maximum training length, we cache and reuse these tensors throughout training.

\subsection{Training Configurations} \label{app:TransferConfig}

We use a fixed block size of 32 and extend each position in the diffusion-based gated convolution layers to at most two future positions. Within this blockwise setting, masking warm-up is applied for GQA layers with sub-block size increasing from 2 to 32 during the second half of the warm-up steps. For the training objectives, we increase $\alpha$ from 0 to 0.5 over the first half of the warm-up steps. Meanwhile, $\rho$ follows the full training schedule: it remains at 0.5 for the first 50\% of total training steps and then gradually increases to 0.8 over the remaining 50\% to encourage higher ground-truth probabilities.

Models are trained for four epochs with a global batch size of 256, using AdamW and a cosine learning-rate schedule initialized at $2 \times 10^{-5}$ for the 1.2B scale and $5 \times 10^{-5}$ for the 350M scale. The first 10\% of training steps are allocated to warm-up. Following standard SFT, the loss is computed only on response tokens. In addition, FlexAttention~\cite{flexattn} is applied to accelerate training with sparse blockwise attention masks. LFM-SFT uses standard padded SFT, while DiCALM uses sequence packing.

\section{Benchmarks and speedup evaluation} \label{app:benchmark}

We evaluate model performance using the `lm-evaluation-harness' library and measure decoding speed separately.\footnote{\url{https://github.com/EleutherAI/lm-evaluation-harness}}

\paragraph{Benchmarks:}
Our experiments employ standard benchmarks, including: MATH500~\cite{hendrycksmath}, using the 4-shot Minerva-style setting~\cite{minerva}; GSM8K~\cite{gsm8k}, using the 5-shot setting; IFBench~\cite{ifbench}, using the 0-shot setting; BBH~\cite{bbh}, using the 0-shot chain-of-thought setting; GPQA-Diamond~\cite{gpqa}, using the 5-shot setting in generative mode; ARC-Challenge~\cite{arc}, using the 0-shot setting in generative mode; and MMLU-Pro~\cite{mmlu_pro}, using the 5-shot setting.

\paragraph{Speed evaluation:}
 Each model is first warmed up using three generation runs on separate input prompts. These warm-up runs serve only to stabilize execution. For each benchmark dataset, decoding speed is then measured on 100 input prompts using a batch size of 1 and a maximum generation length of 1,024 new tokens. All speed experiments are run on one NVIDIA H100 80GB GPU, applying eager attention for GQA layers.  
We time only the "model.generate" call. For each model we compute the throughput (tokens per second), defined as the total number of generated tokens divided by the measured wall-clock time. We then report the speedup, defined as the ratio of our DiCALM's throughput to the LFM-SFT's throughput. We additionally report tokens per forward pass (TPF), computed as the total number of generated tokens divided by the number of model forward passes. Each setting is repeated three times, and we report the average results.

\end{document}